\documentclass[
]{ceurart}

 \usepackage[frozencache=true,cachedir=minted-cache]{minted} 

\usepackage{graphicx}
\usepackage{academicons}
\usepackage{xcolor}
\usepackage{subfig}
\usepackage{rotating}
\usepackage{booktabs} 
\usepackage{subfig}
\usepackage{multirow}

\newif\ifproofread

\newcommand{\changemarker}[1]{%
\ifproofread
\textcolor{red}{#1}%
\else
#1%
\fi
}

\begin{document}
\copyrightyear{2021}
\copyrightclause{Copyright for this paper by its authors.
  Use permitted under Creative Commons License Attribution 4.0
  International (CC BY 4.0).}

\conference{CLEF 2021 -- Conference and Labs of the Evaluation Forum, 
	September 21--24, 2021, Bucharest, Romania}
\proofreadfalse
\title{UPV at CheckThat! 2021: Mitigating Cultural Differences for Identifying Multilingual Check-worthy Claims}
%
%

\author{Ipek Baris Schlicht}[%
email=ibarsch@doctor.upv.es,
]

\author{Angel Felipe Magnossão de Paula}[%
email=adepau@doctor.upv.es,
]

\author{Paolo Rosso}[%
email=prosso@dsic.upv.es,
]

\address{Universitat Politècnica de València, Spain}

\begin{abstract}
Identifying check-worthy claims is \changemarker{often} the first step of automated fact-checking systems. Tackling this task in a multilingual
setting has been understudied. Encoding inputs with multilingual text representations could be one approach to solve the multilingual \changemarker{check-worthiness} detection. However, this approach could suffer if \changemarker{cultural bias exists} within the communities on determining what is check-worthy. In this paper, we propose a language identification task as an auxiliary task to mitigate 
\changemarker{unintended} bias. \changemarker{With this purpose, we experiment joint training by using the datasets from CLEF-2021 CheckThat!, that contain tweets in English, Arabic, Bulgarian, Spanish and Turkish.} Our results show that joint training of language identification and check-worthy claim detection tasks can provide performance gains \changemarker{for some of the selected languages.}

\end{abstract}

\maketitle

\section{Introduction}
The number of fact-checking initiatives worldwide has increased to fight misinformation. Manual fact-checking is a labor-intensive and time-consuming task that \changemarker{cannot} cope up with the dissemination of misinformation~\cite{GravesUnderstanding2018}. Therefore, \changemarker{the} automation of fact-checking steps is required to speed up the process. 

Check-worthy claim detection is a crucial step of an automated fact-checking pipeline~\cite{CazalensLLMT18,GravesUnderstanding2018,ThorneV18} to prioritize what is needed to \changemarker{be fact-checked} by fact-checkers or journalists. There has been \changemarker{an} ongoing effort to address the claim-detection task by \changemarker{different} research communities. Prior studies rely on machine learning methods that use statistical features with bag of words \cite{hassan2015detecting,HassanZACJGHJKN17,gencheva2017context}. Additionally,
CLEF CheckThat! Lab (CTL) has organized shared tasks to tackle this problem in political debates \cite{AtanasovaMBESZK18,AtanasovaNKMM19}  and social media
\cite{barron2020overview}. This year, CTL 2021~\cite{clef-checkthat:2021:LNCS} \changemarker{organized the shared task} in English, Turkish, Bulgarian, Spanish and Arabic \changemarker{where the task} datasets are collected from social media~\cite{clef-checkthat:2021:task1}. The task's input is a tweet and the output \changemarker{is a score indicating the check-worthiness of the tweet.}

Multilingual language models have been widely used in natural language understanding tasks with low-resourced languages (e.g.
comment moderation~\cite{korenvcic2021block}, fake news detection~\cite{Hossain20.1084}). However, the exhibition of cultural differences 
is inevitable in tasks in which cultural context is required~\cite{HwangZLX18}. This issue could harm \changemarker{the transfer of knowledge} across languages. Fact-checking is one of such tasks where disagreements could exist on credibility assessments even among the domain 
experts~\cite{MensioA19,bountouridis2019annotating}. Furthermore, exposure of global claims and their credibility (e.g. Covid-19) could vary by country~\cite{singh2021misinformation}.

With this motivation, in this paper, we present a unified framework that processes the input in different languages and uses a multilingual sentence transformer trained on the mixed language training set to learn representations for the low-resourced languages. To mitigate \changemarker{the bias} in the sentence representations, we introduce a language identification task and train \changemarker{the model} jointly for \changemarker{check-worthiness} detection (CWD) and language identification (LI) tasks.

\changemarker{Our contributions can} be summarized as follows:

\begin{enumerate}
    \item We introduce a framework whose aim is to be aware of cultural bias. We conduct an extensive analysis on its performance.
    \item We employ joint learning to reduce \changemarker{unintended} bias. To the best of our knowledge, a similar method \changemarker{has not} been applied to reduce bias in multilingual fact-checking tasks.
    \item Our framework could be extended with various multilingual transformer models in Huggingface~\cite{wolf-etal-2020-transformers}. \changemarker{The source code} and the trained models are publicly available~\footnote{\url{https://github.com/isspek/Cross_Lingual_Checkworthy_Detection}}. 
\end{enumerate}

\section{Related Work}
ClaimBuster is the first study to address the check-worthy claim detection task. The component of ClaimBuster \cite{hassan2015detecting,HassanZACJGHJKN17} that detects 
check-worthy claims is trained with a Support Vector Machine (SVM) classifier using tf-idf bag of words, named entity types, POS tags, sentiment, and sentence length as a feature set.
\cite{gencheva2017context} proposed a fully connected neural network model trained on claims and their related political debate content. Last year, CTL 
2020~\cite{barron2020overview} \changemarker{organized a shared CWD task} in English and Arabic for the claims in social media. In this shared task, multilingual transformer models performed well on the
Arabic dataset~\cite{hasanain2020bigir}. However, for the English datasets, the participants did not utilize the multilingual transformer model. In our approach, we fine-tune the multilingual sentence transformers~\cite{reimers-2020-multilingual-sentence-bert}, which is computationally less expensive than the BERT models, on the mixed language of the training dataset. We \changemarker{trained one model and employed this for all languages}. 

The multi-task learning approach has been a proven method to mitigate unintended bias. Das et al. \cite{das2018mitigating} applied multi-task learning
on \changemarker{a} face recognition task using Convolution Neural Network. As a related example in the Natural Language Processing (NLP) domain, Vaidya et al. ~\cite{vaidya2020empirical} mitigate the identity bias in toxic comment detection. Their model encodes the inputs with a Bidirectional Long Short-Term Memory Network (BiLSTM). However, our approach and the tasks we deal with are different from those studies.

\section{Methodology}
\begin{figure*}[!ht]
    \centering
    \includegraphics[width=\textwidth]{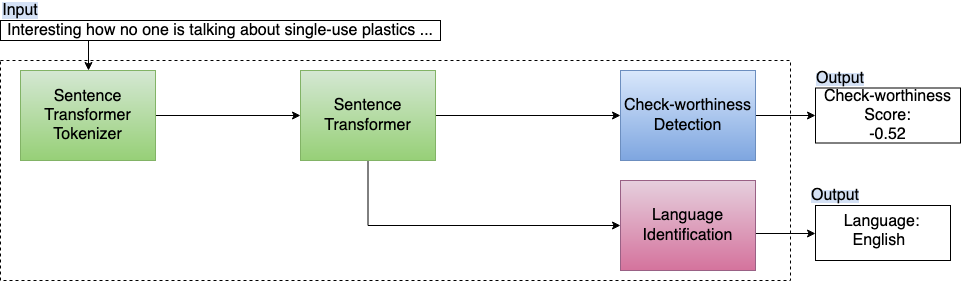}
    \caption{Our proposed framework (QDMSBERT$_{joint}$) for mitigating \changemarker{unintended} bias.}
    \label{fig:architecture}
\end{figure*}

In this section, we introduce our framework, which is depicted in Figure~\ref{fig:architecture}. The input of the framework is a \changemarker{Twitter post}. The input is tokenized with a sentence transformer encoder in
order to \changemarker{be fed} into the transformer layer. After obtaining the shared text representation from a sentence transformer, \changemarker{the framework fine-tunes the shared representation and the classification layers for CWD and LI tasks by minimizing a joint loss.} In the following \changemarker{subsections,} we give more details \changemarker{about} the sentence transformer \changemarker{and the} joint training.

\subsection{Sentence Transformer}
The framework uses a Sentence-BERT (SBERT) transformer~\cite{ReimersG19}, which is a modified BERT that uses a siamese and \changemarker{a} triplet network. 
The SBERT can provide semantically more meaningful sentence embeddings than the BERT models. To support multilingualism in our framework and to enable fine-tuning with a small GPU, we use a pre-trained SBERT that was obtained by applying knowledge distillation~\cite{reimers-2020-multilingual-sentence-bert} and \changemarker{that was} trained on a multilingual corpus from a community-driven Q\&A website\footnote{\url{https://www.quora.com/}}. \changemarker{We refer to it as QDMSBERT.}

We apply mean pooling on the output of  QDMSBERT to obtain sentence embeddings. We set the maximum length of the tokens as 128 by padding shorter texts and truncating longer texts.  

\subsection{Joint Learning}
The framework contains two task layers: one is for the CWD task and the other is for the \changemarker{LI} task.  The input of the task layers \changemarker{are} shared QDMSBERT embeddings. Both task layers use the same neural network structure\changemarker{,} consisting of \changemarker{two fully-connected layers} followed by a softmax layer that outputs the probabilities of task classes. 
During the training, the weighted loss of \changemarker{the} CWD and the \changemarker{LI task} are summed \changemarker{up to compute the} joint loss as seen in Equation~\ref{eq1} where $\alpha$
is a probability indicating the importance of the tasks. Lastly, the joined loss is minimized by optimizing the weights of \changemarker{the transformer network and the tasks' classification layers.}

\begin{equation} \label{eq1}
J_{joint}= \alpha J_{CWD} + (1-\alpha)J_{LI}
\end{equation}

\section{Experiments}

\changemarker{In this section, we give the details of the CLEF 2021 CheckThat! dataset, explain the baselines and the systems that we compared, and present the experimental settings.}

\subsection{Dataset}
The CLEF 2021 CheckThat! offers datasets in English, Spanish, Arabic, Turkish, and Bulgarian for the
CWD task. The statistics of the datasets are given in Table~\ref{tab:data_stats}. The class distribution of the datasets for each language is highly imbalanced, which reflects the real-world cases. Check-worthy samples (Pos-Class) are the minority. English and Bulgarian datasets contain only COVID
topic. The Turkish dataset covers miscellaneous \changemarker{topics}, and the Spanish dataset has \changemarker{only samples about politics.} The topics of the Arabic dataset are mainly COVID related.

\begin{table*}[]
    \centering
    \caption{\changemarker{Topic, class distribution and average tokens in the} CheckThat! dataset. Pos-Class \changemarker{means check-worthy, Neg-Class means not check-worthy.}}
    \label{tab:data_stats}
    \begin{tabular}{{l}*{5}{c}r}
        \toprule
        \textbf{Properties} & \textbf{English} & \textbf{Turkish} & \textbf{Bulgarian} & \textbf{Arabic} & \textbf{Spanish} \\
        \toprule
        \textbf{Topic} & Covid-19 & Miscellaneous & Covid-19 & \changemarker{Miscellaneous} & Politics\\
        \textbf{Pos-Class (Train)} & 290 & 729 & 392 & 921 & 200 \\
        \textbf{Neg-Class (Train)} & 532 & 1170 & 2608 & 2798 & 2295\\
        \textbf{Pos-Class (Dev)} & 60 & 146 & 62 & 107 & 109 \\
        \textbf{Neg-Class (Dev)} & 80 & 242 & 288 & 279 & 1138 \\
        \textbf{Pos-Class (Test)} & 19 & 183 & 76 & 242 & 120\\
        \textbf{Neg-Class (Test)} & 331 & 830 & 281 & 358 & 1128 \\
        \textbf{Avg. Tokens (Train)} & 31.69 & 19.11 & 20.27 & 27.85 & 36.73 \\
        \textbf{Avg. Tokens (Dev)} & 34.71 & 18.22 & 16.66 & 36.68 & 36.19 \\
        \textbf{Avg. Tokens (Test)} & 35.33 & 23.72 & 17.02 & 23.47 & 36.21 \\
        \bottomrule
    \end{tabular}

\end{table*}

\subsection{Baselines}
We compare the proposed model (QDMSBERT$_{joint}$) against the following \changemarker{models and systems:}

\begin{itemize}
    \item \textbf{SVM:} It encodes the texts with unigrams. 
    \item \textbf{Monolingual Models and Mk-Bg-BERT:} We use a distilled SBERT~\cite{ReimersG19} model\footnote{\url{https://huggingface.co/sentence-transformers/distilbert-base-nli-stsb-mean-tokens}} for the English samples. We couldn't find any monolingual SBERTs for Arabic, Turkish and Spanish; therefore, we use popular BERT~\cite{DevlinCLT19} variants that are trained on monolingual \changemarker{corpora}. TrBERT~\footnote{\url{https://huggingface.co/dbmdz/bert-base-turkish-cased}} is the model for Turkish samples, BETO~\cite{CaneteCFP2020} for Spanish, and lastly AraBERT~\cite{antoun2020arabert} for the tweets in Arabic. For \changemarker{Bulgarian tweets}, we leverage a BERT model (Mk-Bg-BERT) trained on Macedonian and Bulgarian \changemarker{corpora}~\footnote{\url{https://huggingface.co/anon-submission-mk/bert-base-macedonian-bulgarian-cased}}. 
    \item \textbf{CLEF-2021:} \changemarker{Submissions for the CLEF-2021 CWD task~\cite{clef-checkthat:2021:task1} that support all languages, namely Accenture, BigIR and TOBB ETU\footnote{At \changemarker{the} time of the writing the paper, we didn't know the system descriptions of their models}}.
    \item \textbf{QDMSBERT:} \changemarker{QDMSBERT$_{joint}$ where the weights are only optimized for the CWD task.} 
\end{itemize}

\subsection{Experimental Settings and Environment}
We split the training dataset randomly into \changemarker{five chunks} and thus train five different QDMSBERT models with the epochs of 3, weighted decay Adam optimizer~\cite{DBLP:conf/iclr/LoshchilovH19}, \changemarker{and in} batches of 16. \changemarker{The mean of each model's predictions represents the final score}. We use the GPU of Google Colab\footnote{\url{https://colab.research.google.com/}} for training the models. \\

\begin{table}[!ht]
    \centering
    \caption{The results of the models on the test set. \changemarker{Our submission is \textbf{QDMSBERT$_{joint}$}.}}
    \label{tab:results_5fold}
    \begin{tabular}{*{2}{l}*{9}{c}r}
        \toprule
        \textbf{Language} & \textbf{Models} & \textbf{MAP} & \textbf{R-Rank} & \textbf{R-Pr} & \textbf{P@1} & \textbf{P@3} & \textbf{P@5} & \textbf{P@10} & \textbf{P@20} & \textbf{P@50} \\
        \toprule
        \multirow{7}{*}{\textbf{English}} & SVM & 0.052	& 0.020	& 0.000	& 0.000	&0.000	&0.000	&0.000	&0.000	&0.020 \\
                & SBERT & \textbf{0.198} & \textbf{1.000} & \textbf{0.211} & \textbf{1.000} & \textbf{0.333} & \textbf{0.200} & \textbf{0.300} & \textbf{0.200} & \textbf{0.160} \\
                \cmidrule{2-11}
                & Accenture & 0.101 & 0.143 & 0.158 & 0.000 & 0.000 & 0.000 & 0.200 & 0.200 & 0.100 \\
                & BigIR & 0.136 & 0.500 & 0.105 & 0.000 & 0.333 & 0.200 & 0.100 & 0.100 & 0.120 \\
                & TOBB ETU & 0.081 & 0.077 & 0.053 & 0.000 & 0.000 & 0.000 & 0.000 & 0.050 & 0.080 \\
                \cmidrule{2-11}
                & QDMSBERT & 0.114 & 0.500 & 0.105 & 0.00 & 0.333 & 0.200 & 0.100 & 0.100 & 0.100  \\
                & \textbf{QDMSBERT$_{joint}$} & 0.149 & \textbf{1.000} & 0.105 & \textbf{1.000} & 0.333 & 0.200 & 0.200 & 0.100 & 0.120 \\
        \midrule
        \multirow{7}{*}{\textbf{Turkish}} & SVM & 0.354 & \textbf{1.000} & 0.311 &\textbf{1.000} & 0.667 & 0.600 & 0.700 & 0.600 & 0.460 \\
                        & TrBERT & 0.563 &\textbf{1.000} & 0.530 &\textbf{1.000} &\textbf{1.000} &\textbf{1.000} & 0.800 & 0.850 & \textbf{0.780} \\
                        \cmidrule{2-11}
                        & Accenture & 0.402 & 0.250 & 0.415 & 0.000 &0.000 & 0.400 & 0.400 & 0.650 & 0.660 \\
                        & BigIR & 0.525 &\textbf{1.000} & 0.503 &\textbf{1.000} &\textbf{1.000} &\textbf{1.000} & 0.800 & 0.700 &0.720 \\
                        & TOBB ETU & \textbf{0.581} & \textbf{1.000} & \textbf{0.585} &\textbf{1.000} &\textbf{1.000} & 0.800 & 0.700 & 0.750 & 0.660 \\
                        \cmidrule{2-11}
                        & QDMSBERT & 0.549 &\textbf{1.000} & 0.579 &\textbf{1.000} & 0.333 & 0.600 & 0.700 & 0.650 & 0.680 \\
                        & \textbf{QDMSBERT$_{joint}$} & 0.517 &\textbf{1.000} & 0.508 &\textbf{1.000} &\textbf{1.000} &\textbf{1.000} &\textbf{1.000} & \textbf{0.850} & 0.700 \\
        \midrule
        \multirow{7}{*}{\textbf{Bulgarian}} & SVM & 0.588	&\textbf{1.000}	&0.474	&\textbf{1.000}	&\textbf{1.000}	&\textbf{1.000}	&0.900&	0.750	&0.640 \\
                        & Mk-Bg-BERT & 0.661 &\textbf{1.000} & 0.645 &\textbf{1.000} &\textbf{1.000} &\textbf{1.000} & 0.900 & 0.700 & 0.700 \\
                         \cmidrule{2-11}
                         & Accenture & 0.497 &\textbf{1.000} & 0.474 &\textbf{1.000} &\textbf{1.000} & 0.800 & 0.700 & 0.600 & 0.440 \\
                         & BigIR & \textbf{0.737} &\textbf{1.000} & 0.632 &\textbf{1.000} &\textbf{1.000} &\textbf{1.000} &\textbf{1.000} &\textbf{1.000} & \textbf{0.800} \\
                         & TOBB ETU & 0.149 & 0.143 & 0.039 & 0.000 & 0.000 & 0.000 & 0.200 & 0.100 & 0.060 \\
                         \cmidrule{2-11}
                         & QDMSBERT & 0.667 &\textbf{1.000} & 0.566 & \textbf{1.000} & \textbf{1.000} & \textbf{1.000} & \textbf{1.000} & 0.900 & 0.720 \\
                         & \textbf{QDMSBERT$_{joint}$} & 0.673 & \textbf{1.000} & 0.605 & \textbf{1.000} & \textbf{1.000} & \textbf{1.000} & \textbf{1.000} & 0.800 & 0.700 \\
        \midrule
        \multirow{7}{*}{\textbf{Arabic}} & SVM &	0.428	&0.500&	0.409&	0.000&	0.667&	0.600&	0.500&	0.450&	0.440	\\
                        & AraBERT & 0.640 &\textbf{1.000} & 0.591 &\textbf{1.000} &\textbf{1.000} & 0.600 & 0.800 & 0.750 & 0.760 \\
                        \cmidrule{2-11}
                        & Accenture & \textbf{0.658} &\textbf{1.000} &	0.599&	\textbf{1.000}&	\textbf{1.000}&	\textbf{1.000}&	\textbf{1.000}&	\textbf{0.950} &	\textbf{0.840} \\
                        & BigIR & 0.615 &	0.500	&0.579	&0.000&	0.667&	0.600&	0.600&	0.800&	0.740 \\
                        & TOBB ETU & 0.575	& 0.333	&0.574	&0.000	&0.333&	0.400&	0.400&	0.500&	0.680 \\
                      \cmidrule{2-11}
                        & QDMSBERT &  0.571 & \textbf{1.000}& 0.579 & 0.000 & 0.667 & 0.600 & 0.600 & 0.550 & 0.580 \\
                        & \textbf{QDMSBERT$_{joint}$ } &  0.548 &\textbf{1.000} & 0.550 & \textbf{1.000}& 0.667 & 0.600 & 0.500 & 0.400 & 0.580 \\
                        
        \midrule
        \multirow{3}{*}{\textbf{Spanish}} & SVM & 0.450 &\textbf{1.000} & 0.450 &\textbf{1.000} & 0.667 & \textbf{0.800} & 0.700 & 0.700 & 0.660\\
                        & BETO & \textbf{0.569} &\textbf{1.000} & \textbf{0.533} &\textbf{1.000} & 0.667 & \textbf{0.800} & 0.800 & \textbf{0.750} & \textbf{0.720} \\
                        \cmidrule{2-11}
                        & Accenture & 0.491 &\textbf{1.000} & 0.508 &\textbf{1.000} & 0.667 & \textbf{0.800} & \textbf{0.900} & 0.700 & 0.620 \\
                        & BigIR & 0.496 &\textbf{1.000} & 0.483 &\textbf{1.000} &\textbf{1.000} & \textbf{0.800} & 0.800 & 0.600 & 0.620 \\
                        & TOBB ETU & 0.537 &\textbf{1.000} & 0.525 &\textbf{1.000} &\textbf{1.000} & \textbf{0.800} & 0.900 & 0.700 & 0.680 \\
                        \cmidrule{2-11}
                        & QDMSBERT & 0.398 & 0.500 & 0.425 & 0.000 & 0.333 & 0.600 & 0.600 & 0.500 & 0.580 \\
                        & \textbf{QDMSBERT$_{joint}$} & 0.446 & 0.333 & 0.475 & 0.000 & 0.333 & 0.600 & 0.800 & 0.650 & 0.580  \\
        \bottomrule
    \end{tabular}

\end{table}

\section{Results}
Table~\ref{tab:results_5fold} presents the results of \changemarker{each model}. We report the test results in official metrics of the \changemarker{shared task: Mean Average Precision (MAP)}, precision scores at 1-50 (P@1-P@50), R-Precision (R-Prec), and R-Rank. We first compare QDMSBERT$_{joint}$ with the SVM and  QDMSBERT. QDMSBERT$_{joint}$ outperforms  QDMSBERT in many metrics across the languages \changemarker{except for Arabic}, also QDMSBERT$_{joint}$ underperforms the SVM in Spanish. We see performance gains on \changemarker{the English, Bulgarian and Turkish samples}. The results indicate that QDMSBERT$_{joint}$ \changemarker{presents better results on the examples in COVID-19, but is generalized less to other topics in Spanish and in Arabic.}

Among the results by the teams who submitted runs in all languages \changemarker{(group CLEF-2021)}, the performance of  QDMSBERT$_{joint}$ is the best in English and the second in Bulgarian which is promising for a low-resource language. 

Monolingual BERT models outperformed our model and the other teams' submissions in English and Spanish.
TrBERT and AraBERT also show better results than our approach. Although we improve our \changemarker{outcome compared to} QDMSBERT by \changemarker{mitigating differences across the languages}, the performance of the monolingual embeddings is still unsurpassed in this task. 

\begin{table}[!ht]
    \centering
    \caption{The performance of  QDMSBERT$_{joint}$ under the different $\alpha$ values}
    \label{tab:results_different_weights}
    \begin{tabular}{*{2}{c}*{8}{l}r}
        \toprule
        \textbf{Language} & \textbf{$\alpha$}  &  \textbf{MAP}  & \textbf{R-Rank} & \textbf{R-Pr} & \textbf{P@1} & \textbf{P@3} & \textbf{P@5} & \textbf{P@10} & \textbf{P@20} & \textbf{P@50} \\
        \toprule
        \multirow{7}{*}{\textbf{English}} & \textbf{0.3}  & 0.143 & \textbf{1.000} & 0.105 & \textbf{1.000} &  \textbf{0.333} & 0.200 & \textbf{0.200} & 0.100 & 0.080 \\
        & \textbf{0.4} & 0.145 & \textbf{1.000} & 0.105 & \textbf{1.000} &  \textbf{0.333} & 0.200 & \textbf{0.200} & 0.100 & 0.080 \\
        & \textbf{0.5}  & 0.151 & \textbf{1.000} & 0.105 & \textbf{1.000} &   \textbf{0.333}& \textbf{0.400} & \textbf{0.200} & 0.100 & 0.080 \\
        & \textbf{0.6} & 0.149 & \textbf{1.000} & 0.105 & \textbf{1.000} &   \textbf{0.333} & 0.200 & \textbf{0.200} & 0.100 & 0.120 \\
        & \textbf{0.7} & 0.123 & 0.500 & 0.105 & 0.00 &   \textbf{0.333} & 0.200 & \textbf{0.200} & 0.100 & 0.120\\
        & \textbf{0.8} & \textbf{0.155} & \textbf{1.000} & \textbf{0.158} & \textbf{1.000} &   \textbf{0.333} & 0.200 & \textbf{0.200} & \textbf{0.150} & \textbf{0.120} \\
        & \textbf{0.9} & 0.144 & \textbf{1.000} & 0.105 & \textbf{1.000} &   \textbf{0.333} & 0.200 & 0.100 & \textbf{0.150} & \textbf{0.120}\\
        \midrule
        \multirow{7}{*}{\textbf{Turkish}} & \textbf{0.3} & 0.520 & \textbf{1.000} & 0.492 & \textbf{1.000} & \textbf{1.000} & \textbf{1.000} & \textbf{1.000} & 0.900 & 0.660 \\
        & \textbf{0.4}  & 0.531 & \textbf{1.000} & 0.481 & \textbf{1.000} & \textbf{1.000} & \textbf{1.000} & \textbf{1.000} & 0.950 & 0.740 \\
        & \textbf{0.5}  & 0.534 & \textbf{1.000} & 0.492 & \textbf{1.000} & \textbf{1.000} & \textbf{1.000} & \textbf{1.000} & 0.950 & 0.740  \\
        & \textbf{0.6}  & 0.517 & \textbf{1.000} & 0.508 & \textbf{1.000} & \textbf{1.000} & \textbf{1.000} & \textbf{1.000} & 0.850 & 0.700 \\
        & \textbf{0.7} & 0.528 & \textbf{1.000} & 0.497 & \textbf{1.000} & \textbf{1.000} & 0.200 & 0.200 & 0.850 & 0.680 \\
        & \textbf{0.8}  & \textbf{0.588} & \textbf{1.000} & 0.563 & \textbf{1.000} & \textbf{1.000} & \textbf{1.000} & \textbf{1.000} & \textbf{0.950} & \textbf{0.780} \\
        & \textbf{0.9}  & 0.582 & \textbf{1.000} & \textbf{0.568} & \textbf{1.000} & \textbf{1.000} & \textbf{1.000} & \textbf{1.000} & 0.850 & 0.740 \\

        \midrule
        \multirow{7}{*}{\textbf{Bulgarian}} & \textbf{0.3}  & 0.657 & \textbf{1.000} & 0.618 & \textbf{1.000} & \textbf{1.000} & \textbf{1.000} & 0.900 & 0.800 & \textbf{0.720} \\
        & \textbf{0.4} & 0.666 & \textbf{1.000} & 0.618 & \textbf{1.000} & \textbf{1.000} & \textbf{1.000} & 0.900 & 0.800 & 0.700  \\
        & \textbf{0.5}  & 0.670 & \textbf{1.000} & 0.618 & \textbf{1.000} & \textbf{1.000} & \textbf{1.000} & 0.900 & 0.850 & \textbf{0.720} \\
        & \textbf{0.6} & 0.673 & \textbf{1.000} & 0.605 & \textbf{1.000} & \textbf{1.000} & \textbf{1.000} & \textbf{1.000} & 0.800 & 0.700  \\
        & \textbf{0.7}  & \textbf{0.677} & \textbf{1.000} & 0.618 & \textbf{1.000} & \textbf{1.000} & \textbf{1.000} & \textbf{1.000} & \textbf{0.850} & 0.700 \\
        & \textbf{0.8} & 0.670 & \textbf{1.000} & 0.592 & \textbf{1.000} & \textbf{1.000} & \textbf{1.000} & \textbf{1.000} & 0.800 & \textbf{0.720} \\
        & \textbf{0.9}  &\textbf{0.677} & \textbf{1.000} & 0.579 & \textbf{1.000} & \textbf{1.000} & \textbf{1.000} & 0.900 & \textbf{0.850} & 0.700  \\
                        
        \midrule
        \multirow{7}{*}{\textbf{Arabic}} & \textbf{0.3}  & 0.562 & \textbf{1.000} & 0.558 & \textbf{1.000} & 0.333 & 0.400 & \textbf{0.500} & 0.400 & \textbf{0.680} \\
        & \textbf{0.4} & 0.561 & \textbf{1.000} & 0.562 & \textbf{1.000} & \textbf{0.667} & 0.400 & 0.400 & 0.350 & 0.640  \\ 
        & \textbf{0.5}  & 0.567 & \textbf{1.000} & 0.562 & \textbf{1.000} & \textbf{0.667} & 0.400 & 0.400 & 0.400 & 0.660 \\
        & \textbf{0.6} &  0.548 & \textbf{1.000} & 0.550 & \textbf{1.000} & \textbf{0.667} & \textbf{0.600} & \textbf{0.500} & 0.400 & 0.580 \\
         & \textbf{0.7}  & 0.561 & \textbf{1.000} & 0.566 & \textbf{1.000} & \textbf{0.667} & 0.400 & \textbf{0.500} & 0.400 & 0.620  \\
         & \textbf{0.8}  & 0.566 & \textbf{1.000} & 0.566 & \textbf{1.000} & \textbf{0.667} & 0.400 & 0.400 & 0.450 & 0.580 \\
         & \textbf{0.9} & \textbf{0.573} & \textbf{1.000} & \textbf{0.574} & \textbf{1.000} & \textbf{0.667} & 0.400 & \textbf{0.500} & \textbf{0.500} & 0.580  \\
         
        \midrule
        \multirow{7}{*}{\textbf{Spanish}} & \textbf{0.3} & 0.450 & \textbf{0.333} & 0.458 & 0.00 & \textbf{0.333} & \textbf{0.600} & 0.700 & \textbf{0.750} & 0.580\\
        & \textbf{0.4}  & 0.453 & \textbf{0.333} & 0.475 & 0.00 & \textbf{0.333} & \textbf{0.600}& 0.700 & \textbf{0.750} & 0.580  \\
        & \textbf{0.5} & \textbf{0.456} & \textbf{0.333} & 0.472 & 0.00 & \textbf{0.333} &\textbf{0.600} & 0.700 & \textbf{0.750} & \textbf{0.640} \\
        & \textbf{0.6} & 0.446 & \textbf{0.333} & 0.475 & 0.00 & \textbf{0.333}& \textbf{0.600} & \textbf{0.800} & 0.650 & 0.580 \\
        & \textbf{0.7}  & 0.443 & \textbf{0.333} & \textbf{0.483} & 0.00 & \textbf{0.333} & 0.400 & \textbf{0.600} & 0.600 & 0.580 \\
        & \textbf{0.8} & 0.443 & \textbf{0.333} & 0.475 & 0.00 & \textbf{0.333} & 0.400 & 0.500 & 0.650 & 0.580 \\
        & \textbf{0.9} & 0.431 & 0.250 & 0.467 & 0.00 & 0.00 & 0.400 & 0.500 & 0.700 & 0.580  \\
         
    \bottomrule
    \end{tabular}

\end{table}  

\changemarker{The presented results of QDMSBERT$_{joint}$ were accomplished by using a contribution of task loss ($\alpha$) of 0.6.
This initial value was choose heuristically.}
As an ablation study, \changemarker{we change the $\alpha$ values of the tasks' loss and train  QDMSBERT$_{joint}$ for each $\alpha$ value to understand its influence on the CWD learning. The optimal alpha value is 0.8.} In Bulgarian samples, the lower alpha values could also yield good performance for the CWD task. 

\begin{figure}
\centering
\includegraphics[width=0.8\textwidth]{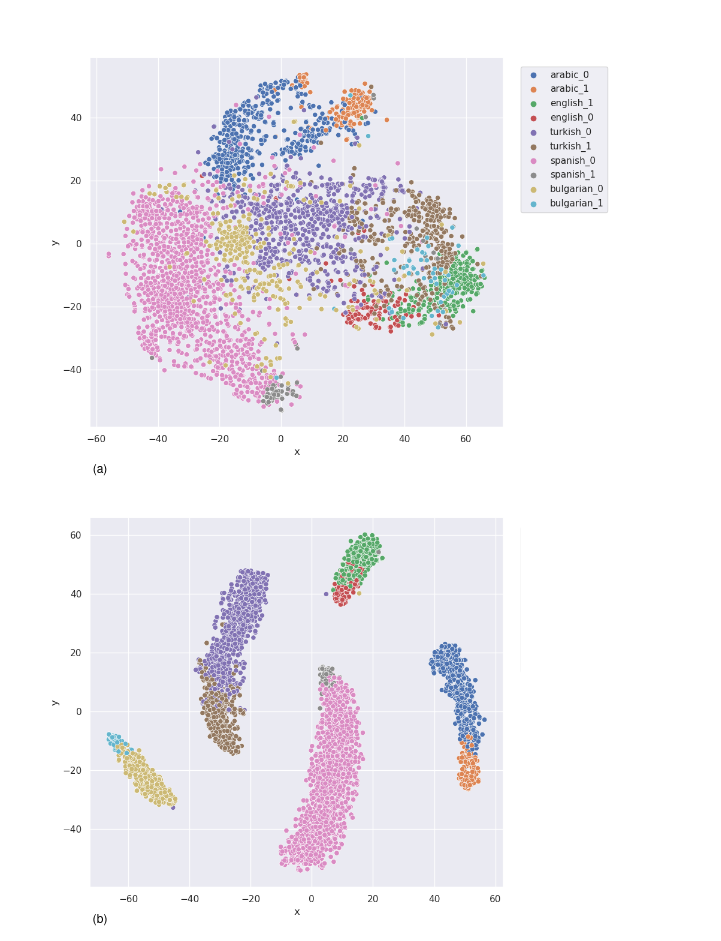}
\caption{T-SNE visualization of  QDMSBERT (a) and QDMSBERT$_{joint}$ (b). 0: not check-worthy, 1: check-worthy}
\label{t-sne}
\end{figure}

Lastly, we analyze the feature representations of  QDMSBERT and QDMSBERT$_{joint}$. We visualize the feature representations by applying t-distributed stochastic neighbor embedding nonlinear dimensionality reduction (T-SNE)~\cite{van2008visualizing}.
As depicted in Figure~\ref{t-sne}, the features that QDMSBERT$_{joint}$ produces are more clearly separated.
\changemarker{For instance, the cluster with English samples (lower right region of Figure~\ref{t-sne}a) in the T-SNE for QDMSBERT overlaps with both the cluster of Turkish and the cluster of Bulgarian samples.
In contrast, the T-SNE for QDMSBERT$_{joint}$ shows that only very few non-English samples fall close to the English cluster (upper region of Figure~\ref{t-sne}b).}

\section{Conclusion}
In this paper, we proposed a method to tackle the multilingual check-worthiness of claims. To mitigate bias due to \changemarker{cultural differences,} we leveraged multilingual sentence BERTs as feature representations and trained \changemarker{them} jointly with the language identification task. Our approach outperformed the SVM and QDMSBERT \changemarker{for almost all of the languages} on the CLEF2021 dataset. Also, it became one of the 
top-performing approaches in Bulgarian and English datasets among the submissions that have been done for \changemarker{all these} languages. In the future, we will \changemarker{investigate how the consideration of images \cite{DBLP:conf/www/CheemaHME21} that are embedded in the tweets influence the results.}

\section*{Acknowledgement}
The work of P. Rosso was partially funded by the Spanish Ministry of
Science and Innovation under the research project MISMIS-FAKEnHATE on
MISinformation and MIScommunication in social media: FAKE news and HATE
speech (PGC2018-096212-B-C31).

\bibliography{paper}

\begin{thebibliography}{29}
\expandafter\ifx\csname natexlab\endcsname\relax\def\natexlab#1{#1}\fi
\providecommand{\url}[1]{\texttt{#1}}
\providecommand{\href}[2]{#2}
\providecommand{\path}[1]{#1}
\providecommand{\DOIprefix}{doi:}
\providecommand{\ArXivprefix}{arXiv:}
\providecommand{\URLprefix}{URL: }
\providecommand{\Pubmedprefix}{pmid:}
\providecommand{\doi}[1]{\href{http://dx.doi.org/#1}{\path{#1}}}
\providecommand{\Pubmed}[1]{\href{pmid:#1}{\path{#1}}}
\providecommand{\bibinfo}[2]{#2}
\ifx\xfnm\relax \def\xfnm[#1]{\unskip,\space#1}\fi
\bibitem[{Graves(2018)}]{GravesUnderstanding2018}
\bibinfo{author}{L.~Graves},
\newblock \bibinfo{title}{Understanding the promise and limits of automated
  fact-checking},
\newblock \bibinfo{journal}{Factsheet} \bibinfo{volume}{2}
  (\bibinfo{year}{2018}) \bibinfo{pages}{2018--02}.
\bibitem[{Cazalens et~al.(2018)Cazalens, Lamarre, Leblay, Manolescu, and
  Tannier}]{CazalensLLMT18}
\bibinfo{author}{S.~Cazalens}, \bibinfo{author}{P.~Lamarre},
  \bibinfo{author}{J.~Leblay}, \bibinfo{author}{I.~Manolescu},
  \bibinfo{author}{X.~Tannier},
\newblock \bibinfo{title}{A content management perspective on fact-checking},
\newblock in: \bibinfo{booktitle}{{WWW} (Companion Volume)},
  \bibinfo{publisher}{{ACM}}, \bibinfo{year}{2018}, pp.
  \bibinfo{pages}{565--574}.
\bibitem[{Thorne and Vlachos(2018)}]{ThorneV18}
\bibinfo{author}{J.~Thorne}, \bibinfo{author}{A.~Vlachos},
\newblock \bibinfo{title}{Automated fact checking: Task formulations, methods
  and future directions},
\newblock in: \bibinfo{booktitle}{{COLING}}, \bibinfo{publisher}{Association
  for Computational Linguistics}, \bibinfo{year}{2018}, pp.
  \bibinfo{pages}{3346--3359}.
\bibitem[{Hassan et~al.(2015)Hassan, Li, and Tremayne}]{hassan2015detecting}
\bibinfo{author}{N.~Hassan}, \bibinfo{author}{C.~Li},
  \bibinfo{author}{M.~Tremayne},
\newblock \bibinfo{title}{Detecting check-worthy factual claims in presidential
  debates},
\newblock in: \bibinfo{booktitle}{Proceedings of the 24th acm international on
  conference on information and knowledge management}, \bibinfo{year}{2015},
  pp. \bibinfo{pages}{1835--1838}.
\bibitem[{Hassan et~al.(2017)Hassan, Zhang, Arslan, Caraballo, Jimenez,
  Gawsane, Hasan, Joseph, Kulkarni, Nayak, Sable, Li, and
  Tremayne}]{HassanZACJGHJKN17}
\bibinfo{author}{N.~Hassan}, \bibinfo{author}{G.~Zhang},
  \bibinfo{author}{F.~Arslan}, \bibinfo{author}{J.~Caraballo},
  \bibinfo{author}{D.~Jimenez}, \bibinfo{author}{S.~Gawsane},
  \bibinfo{author}{S.~Hasan}, \bibinfo{author}{M.~Joseph},
  \bibinfo{author}{A.~Kulkarni}, \bibinfo{author}{A.~K. Nayak},
  \bibinfo{author}{V.~Sable}, \bibinfo{author}{C.~Li},
  \bibinfo{author}{M.~Tremayne},
\newblock \bibinfo{title}{Claimbuster: The first-ever end-to-end fact-checking
  system},
\newblock \bibinfo{journal}{Proc. {VLDB} Endow.} \bibinfo{volume}{10}
  (\bibinfo{year}{2017}) \bibinfo{pages}{1945--1948}.
\bibitem[{Gencheva et~al.(2017)Gencheva, Nakov, M{\`a}rquez,
  Barr{\'o}n-Cede{\~n}o, and Koychev}]{gencheva2017context}
\bibinfo{author}{P.~Gencheva}, \bibinfo{author}{P.~Nakov},
  \bibinfo{author}{L.~M{\`a}rquez}, \bibinfo{author}{A.~Barr{\'o}n-Cede{\~n}o},
  \bibinfo{author}{I.~Koychev},
\newblock \bibinfo{title}{A context-aware approach for detecting worth-checking
  claims in political debates},
\newblock in: \bibinfo{booktitle}{Proceedings of the International Conference
  Recent Advances in Natural Language Processing, RANLP 2017},
  \bibinfo{year}{2017}, pp. \bibinfo{pages}{267--276}.
\bibitem[{Atanasova et~al.(2018)Atanasova, M{\`{a}}rquez,
  Barr{\'{o}}n{-}Cede{\~{n}}o, Elsayed, Suwaileh, Zaghouani, Kyuchukov,
  Martino, and Nakov}]{AtanasovaMBESZK18}
\bibinfo{author}{P.~Atanasova}, \bibinfo{author}{L.~M{\`{a}}rquez},
  \bibinfo{author}{A.~Barr{\'{o}}n{-}Cede{\~{n}}o},
  \bibinfo{author}{T.~Elsayed}, \bibinfo{author}{R.~Suwaileh},
  \bibinfo{author}{W.~Zaghouani}, \bibinfo{author}{S.~Kyuchukov},
  \bibinfo{author}{G.~D.~S. Martino}, \bibinfo{author}{P.~Nakov},
\newblock \bibinfo{title}{Overview of the {CLEF-2018} checkthat! lab on
  automatic identification and verification of political claims. task 1:
  Check-worthiness},
\newblock in: \bibinfo{booktitle}{{CLEF} (Working Notes)}, volume
  \bibinfo{volume}{2125} of \textit{\bibinfo{series}{{CEUR} Workshop
  Proceedings}}, \bibinfo{publisher}{CEUR-WS.org}, \bibinfo{year}{2018}.
\bibitem[{Atanasova et~al.(2019)Atanasova, Nakov, Karadzhov, Mohtarami, and
  Martino}]{AtanasovaNKMM19}
\bibinfo{author}{P.~Atanasova}, \bibinfo{author}{P.~Nakov},
  \bibinfo{author}{G.~Karadzhov}, \bibinfo{author}{M.~Mohtarami},
  \bibinfo{author}{G.~D.~S. Martino},
\newblock \bibinfo{title}{Overview of the {CLEF-2019} checkthat! lab: Automatic
  identification and verification of claims. task 1: Check-worthiness},
\newblock in: \bibinfo{booktitle}{{CLEF} (Working Notes)}, volume
  \bibinfo{volume}{2380} of \textit{\bibinfo{series}{{CEUR} Workshop
  Proceedings}}, \bibinfo{publisher}{CEUR-WS.org}, \bibinfo{year}{2019}.
\bibitem[{Barr{\'{o}}n{-}Cede{\~{n}}o et~al.(2020)Barr{\'{o}}n{-}Cede{\~{n}}o,
  Elsayed, Nakov, Martino, Hasanain, Suwaileh, Haouari, Babulkov, Hamdan,
  Nikolov, Shaar, and Ali}]{barron2020overview}
\bibinfo{author}{A.~Barr{\'{o}}n{-}Cede{\~{n}}o}, \bibinfo{author}{T.~Elsayed},
  \bibinfo{author}{P.~Nakov}, \bibinfo{author}{G.~D.~S. Martino},
  \bibinfo{author}{M.~Hasanain}, \bibinfo{author}{R.~Suwaileh},
  \bibinfo{author}{F.~Haouari}, \bibinfo{author}{N.~Babulkov},
  \bibinfo{author}{B.~Hamdan}, \bibinfo{author}{A.~Nikolov},
  \bibinfo{author}{S.~Shaar}, \bibinfo{author}{Z.~S. Ali},
\newblock \bibinfo{title}{Overview of checkthat! 2020: Automatic identification
  and verification of claims in social media},
\newblock in: \bibinfo{booktitle}{{CLEF}}, volume \bibinfo{volume}{12260} of
  \textit{\bibinfo{series}{Lecture Notes in Computer Science}},
  \bibinfo{publisher}{Springer}, \bibinfo{year}{2020}, pp.
  \bibinfo{pages}{215--236}.
\bibitem[{Nakov et~al.(2021)Nakov, Da~San~Martino, Elsayed,
  Barr{\'{o}}n{-}Cede{\~{n}}o, M\'{i}guez, Shaar, Alam, Haouari, Hasanain,
  Mansour, Hamdan, Ali, Babulkov, Nikolov, Shahi, Struß, Mandl, Kutlu, and
  Kartal}]{clef-checkthat:2021:LNCS}
\bibinfo{author}{P.~Nakov}, \bibinfo{author}{G.~Da~San~Martino},
  \bibinfo{author}{T.~Elsayed},
  \bibinfo{author}{A.~Barr{\'{o}}n{-}Cede{\~{n}}o},
  \bibinfo{author}{R.~M\'{i}guez}, \bibinfo{author}{S.~Shaar},
  \bibinfo{author}{F.~Alam}, \bibinfo{author}{F.~Haouari},
  \bibinfo{author}{M.~Hasanain}, \bibinfo{author}{W.~Mansour},
  \bibinfo{author}{B.~Hamdan}, \bibinfo{author}{Z.~S. Ali},
  \bibinfo{author}{N.~Babulkov}, \bibinfo{author}{A.~Nikolov},
  \bibinfo{author}{G.~K. Shahi}, \bibinfo{author}{J.~M. Struß},
  \bibinfo{author}{T.~Mandl}, \bibinfo{author}{M.~Kutlu},
  \bibinfo{author}{Y.~S. Kartal},
\newblock \bibinfo{title}{Overview of the {CLEF}-2021 {CheckThat}! lab on
  detecting check-worthy claims, previously fact-checked claims, and fake
  news},
\newblock in: \bibinfo{booktitle}{Proceedings of the 12th International
  Conference of the CLEF Association: Information Access Evaluation Meets
  Multiliguality, Multimodality, and Visualization}, CLEF~'2021,
  \bibinfo{address}{Bucharest, Romania (online)}, \bibinfo{year}{2021}.
\bibitem[{Shaar et~al.(2021)Shaar, Hasanain, Hamdan, Ali, Haouari,
  Alex~Nikolov, Yavuz Selim~Kartal, Da~San~Martino,
  Barr{\'{o}}n{-}Cede{\~{n}}o, M\'{i}guez, Elsayed, and
  Nakov}]{clef-checkthat:2021:task1}
\bibinfo{author}{S.~Shaar}, \bibinfo{author}{M.~Hasanain},
  \bibinfo{author}{B.~Hamdan}, \bibinfo{author}{Z.~S. Ali},
  \bibinfo{author}{F.~Haouari}, \bibinfo{author}{M.~K. Alex~Nikolov},
  \bibinfo{author}{F.~A. Yavuz Selim~Kartal},
  \bibinfo{author}{G.~Da~San~Martino},
  \bibinfo{author}{A.~Barr{\'{o}}n{-}Cede{\~{n}}o},
  \bibinfo{author}{R.~M\'{i}guez}, \bibinfo{author}{T.~Elsayed},
  \bibinfo{author}{P.~Nakov},
\newblock \bibinfo{title}{Overview of the {CLEF}-2021 {CheckThat}! lab task 1
  on check-worthiness estimation in tweets and political debates},
\newblock in: \bibinfo{booktitle}{Working Notes of CLEF 2021---Conference and
  Labs of the Evaluation Forum}, CLEF~'2021, \bibinfo{address}{Bucharest,
  Romania (online)}, \bibinfo{year}{2021}.
\bibitem[{Koren{\v{c}}i{\'c} et~al.(2021)Koren{\v{c}}i{\'c}, Baris, Fernandez,
  Leuschel, and Salido}]{korenvcic2021block}
\bibinfo{author}{D.~Koren{\v{c}}i{\'c}}, \bibinfo{author}{I.~Baris},
  \bibinfo{author}{E.~Fernandez}, \bibinfo{author}{K.~Leuschel},
  \bibinfo{author}{E.~Salido},
\newblock \bibinfo{title}{To block or not to block: Experiments with machine
  learning for news comment moderation},
\newblock in: \bibinfo{booktitle}{Proceedings of the EACL Hackashop on News
  Media Content Analysis and Automated Report Generation},
  \bibinfo{year}{2021}, pp. \bibinfo{pages}{127--133}.
\bibitem[{Hossain et~al.(2020)Hossain, Rahman, Islam, and Kar}]{Hossain20.1084}
\bibinfo{author}{M.~Z. Hossain}, \bibinfo{author}{M.~A. Rahman},
  \bibinfo{author}{M.~S. Islam}, \bibinfo{author}{S.~Kar},
\newblock \bibinfo{title}{{BanFakeNews: A Dataset for Detecting Fake News in
  Bangla}},
\newblock in: \bibinfo{booktitle}{Proceedings of the Twelfth International
  Conference on Language Resources and Evaluation (LREC 2020)},
  \bibinfo{publisher}{European Language Resources Association (ELRA)},
  \bibinfo{year}{2020}.
\bibitem[{Lin et~al.(2018)Lin, Xu, Zhu, and Hwang}]{HwangZLX18}
\bibinfo{author}{B.~Y. Lin}, \bibinfo{author}{F.~F. Xu}, \bibinfo{author}{K.~Q.
  Zhu}, \bibinfo{author}{S.~Hwang},
\newblock \bibinfo{title}{Mining cross-cultural differences and similarities in
  social media},
\newblock in: \bibinfo{booktitle}{{ACL} {(1)}}, \bibinfo{publisher}{Association
  for Computational Linguistics}, \bibinfo{year}{2018}, pp.
  \bibinfo{pages}{709--719}.
\bibitem[{Mensio and Alani(2019)}]{MensioA19}
\bibinfo{author}{M.~Mensio}, \bibinfo{author}{H.~Alani},
\newblock \bibinfo{title}{News source credibility in the eyes of different
  assessors},
\newblock in: \bibinfo{booktitle}{{TTO}}, \bibinfo{year}{2019}.
\bibitem[{Bountouridis et~al.(2019)Bountouridis, Makhortykh, Sullivan,
  Harambam, Tintarev, and Hauff}]{bountouridis2019annotating}
\bibinfo{author}{D.~Bountouridis}, \bibinfo{author}{M.~Makhortykh},
  \bibinfo{author}{E.~Sullivan}, \bibinfo{author}{J.~Harambam},
  \bibinfo{author}{N.~Tintarev}, \bibinfo{author}{C.~Hauff},
\newblock \bibinfo{title}{Annotating credibility: Identifying and mitigating
  bias in credibility datasets}  (\bibinfo{year}{2019}).
\bibitem[{Singh et~al.(2021)Singh, Lima, Cha, Cha, Kulshrestha, Ahn, and
  Varol}]{singh2021misinformation}
\bibinfo{author}{K.~Singh}, \bibinfo{author}{G.~Lima},
  \bibinfo{author}{M.~Cha}, \bibinfo{author}{C.~Cha},
  \bibinfo{author}{J.~Kulshrestha}, \bibinfo{author}{Y.-Y. Ahn},
  \bibinfo{author}{O.~Varol},
\newblock \bibinfo{title}{Misinformation, believability, and vaccine acceptance
  over 40 countries: Takeaways from the initial phase of the covid-19
  infodemic},
\newblock \bibinfo{journal}{arXiv preprint arXiv:2104.10864}
  (\bibinfo{year}{2021}).
\bibitem[{Wolf et~al.(2020)Wolf, Debut, Sanh, Chaumond, Delangue, Moi, Cistac,
  Rault, Louf, Funtowicz, Davison, Shleifer, von Platen, Ma, Jernite, Plu, Xu,
  Scao, Gugger, Drame, Lhoest, and Rush}]{wolf-etal-2020-transformers}
\bibinfo{author}{T.~Wolf}, \bibinfo{author}{L.~Debut},
  \bibinfo{author}{V.~Sanh}, \bibinfo{author}{J.~Chaumond},
  \bibinfo{author}{C.~Delangue}, \bibinfo{author}{A.~Moi},
  \bibinfo{author}{P.~Cistac}, \bibinfo{author}{T.~Rault},
  \bibinfo{author}{R.~Louf}, \bibinfo{author}{M.~Funtowicz},
  \bibinfo{author}{J.~Davison}, \bibinfo{author}{S.~Shleifer},
  \bibinfo{author}{P.~von Platen}, \bibinfo{author}{C.~Ma},
  \bibinfo{author}{Y.~Jernite}, \bibinfo{author}{J.~Plu},
  \bibinfo{author}{C.~Xu}, \bibinfo{author}{T.~L. Scao},
  \bibinfo{author}{S.~Gugger}, \bibinfo{author}{M.~Drame},
  \bibinfo{author}{Q.~Lhoest}, \bibinfo{author}{A.~M. Rush},
\newblock \bibinfo{title}{Transformers: State-of-the-art natural language
  processing},
\newblock in: \bibinfo{booktitle}{Proceedings of the 2020 Conference on
  Empirical Methods in Natural Language Processing: System Demonstrations},
  \bibinfo{publisher}{Association for Computational Linguistics},
  \bibinfo{address}{Online}, \bibinfo{year}{2020}, pp. \bibinfo{pages}{38--45}.
  \URLprefix \url{https://www.aclweb.org/anthology/2020.emnlp-demos.6}.
\bibitem[{Hasanain and Elsayed(2020)}]{hasanain2020bigir}
\bibinfo{author}{M.~Hasanain}, \bibinfo{author}{T.~Elsayed},
\newblock \bibinfo{title}{bigir at checkthat! 2020: Multilingual bert for
  ranking arabic tweets by check-worthiness},
\newblock \bibinfo{journal}{Cappellato et al.[10]}  (\bibinfo{year}{2020}).
\bibitem[{Reimers and Gurevych(2020)}]{reimers-2020-multilingual-sentence-bert}
\bibinfo{author}{N.~Reimers}, \bibinfo{author}{I.~Gurevych},
\newblock \bibinfo{title}{Making monolingual sentence embeddings multilingual
  using knowledge distillation},
\newblock in: \bibinfo{booktitle}{Proceedings of the 2020 Conference on
  Empirical Methods in Natural Language Processing},
  \bibinfo{publisher}{Association for Computational Linguistics},
  \bibinfo{year}{2020}.
\bibitem[{Das et~al.(2018)Das, Dantcheva, and Bremond}]{das2018mitigating}
\bibinfo{author}{A.~Das}, \bibinfo{author}{A.~Dantcheva},
  \bibinfo{author}{F.~Bremond},
\newblock \bibinfo{title}{Mitigating bias in gender, age and ethnicity
  classification: a multi-task convolution neural network approach},
\newblock in: \bibinfo{booktitle}{Proceedings of the European Conference on
  Computer Vision (ECCV) Workshops}, \bibinfo{year}{2018}, pp.
  \bibinfo{pages}{0--0}.
\bibitem[{Vaidya et~al.(2020)Vaidya, Mai, and Ning}]{vaidya2020empirical}
\bibinfo{author}{A.~Vaidya}, \bibinfo{author}{F.~Mai},
  \bibinfo{author}{Y.~Ning},
\newblock \bibinfo{title}{Empirical analysis of multi-task learning for
  reducing identity bias in toxic comment detection},
\newblock in: \bibinfo{booktitle}{Proceedings of the International AAAI
  Conference on Web and Social Media}, volume~\bibinfo{volume}{14},
  \bibinfo{year}{2020}, pp. \bibinfo{pages}{683--693}.
\bibitem[{Reimers and Gurevych(2019)}]{ReimersG19}
\bibinfo{author}{N.~Reimers}, \bibinfo{author}{I.~Gurevych},
\newblock \bibinfo{title}{Sentence-bert: Sentence embeddings using siamese
  bert-networks},
\newblock in: \bibinfo{booktitle}{Proceedings of the 2019 Conference on
  Empirical Methods in Natural Language Processing},
  \bibinfo{publisher}{Association for Computational Linguistics},
  \bibinfo{year}{2019}.
\bibitem[{Devlin et~al.(2019)Devlin, Chang, Lee, and Toutanova}]{DevlinCLT19}
\bibinfo{author}{J.~Devlin}, \bibinfo{author}{M.~Chang},
  \bibinfo{author}{K.~Lee}, \bibinfo{author}{K.~Toutanova},
\newblock \bibinfo{title}{{BERT:} pre-training of deep bidirectional
  transformers for language understanding},
\newblock in: \bibinfo{booktitle}{{NAACL-HLT} {(1)}},
  \bibinfo{publisher}{Association for Computational Linguistics},
  \bibinfo{year}{2019}, pp. \bibinfo{pages}{4171--4186}.
\bibitem[{Cañete et~al.(2020)Cañete, Chaperon, Fuentes, Ho, Kang, and
  Pérez}]{CaneteCFP2020}
\bibinfo{author}{J.~Cañete}, \bibinfo{author}{G.~Chaperon},
  \bibinfo{author}{R.~Fuentes}, \bibinfo{author}{J.-H. Ho},
  \bibinfo{author}{H.~Kang}, \bibinfo{author}{J.~Pérez},
\newblock \bibinfo{title}{Spanish pre-trained bert model and evaluation data},
\newblock in: \bibinfo{booktitle}{PML4DC at ICLR 2020}, \bibinfo{year}{2020}.
\bibitem[{Antoun et~al.(????)Antoun, Baly, and Hajj}]{antoun2020arabert}
\bibinfo{author}{W.~Antoun}, \bibinfo{author}{F.~Baly},
  \bibinfo{author}{H.~Hajj},
\newblock \bibinfo{title}{Arabert: Transformer-based model for arabic language
  understanding},
\newblock in: \bibinfo{booktitle}{LREC 2020 Workshop Language Resources and
  Evaluation Conference 11--16 May 2020}, ????, p.~\bibinfo{pages}{9}.
\bibitem[{Loshchilov and Hutter(2019)}]{DBLP:conf/iclr/LoshchilovH19}
\bibinfo{author}{I.~Loshchilov}, \bibinfo{author}{F.~Hutter},
\newblock \bibinfo{title}{Decoupled weight decay regularization},
\newblock in: \bibinfo{booktitle}{{ICLR} (Poster)},
  \bibinfo{publisher}{OpenReview.net}, \bibinfo{year}{2019}.
\bibitem[{Van~der Maaten and Hinton(2008)}]{van2008visualizing}
\bibinfo{author}{L.~Van~der Maaten}, \bibinfo{author}{G.~Hinton},
\newblock \bibinfo{title}{Visualizing data using t-sne.},
\newblock \bibinfo{journal}{Journal of machine learning research}
  \bibinfo{volume}{9} (\bibinfo{year}{2008}).
\bibitem[{Cheema et~al.(2021)Cheema, Hakimov, M{\"{u}}ller{-}Budack, and
  Ewerth}]{DBLP:conf/www/CheemaHME21}
\bibinfo{author}{G.~S. Cheema}, \bibinfo{author}{S.~Hakimov},
  \bibinfo{author}{E.~M{\"{u}}ller{-}Budack}, \bibinfo{author}{R.~Ewerth},
\newblock \bibinfo{title}{On the role of images for analyzing claims in social
  media},
\newblock in: \bibinfo{booktitle}{CLEOPATRA@WWW}, volume \bibinfo{volume}{2829}
  of \textit{\bibinfo{series}{{CEUR} Workshop Proceedings}},
  \bibinfo{publisher}{CEUR-WS.org}, \bibinfo{year}{2021}, pp.
  \bibinfo{pages}{32--46}.

\end{thebibliography}

\end{document}